# Using Deep Q-Learning to Dynamically Toggle between Push/Pull Actions in Computational Trust Mechanisms

**Zoi Lygizou**
Hellenic Open University, Patra, Greece
std084140@ac.eap.gr (Corresponding author)
https://orcid.org/0000-0001-8414-7963

**Dimitris Kalles**
Hellenic Open University, Patra, Greece
kalles@eap.gr (Corresponding author)
https://orcid.org/0000-0003-0364-5966

**Abstract**

Recent work on decentralized computational trust models for open Multi Agent Systems has resulted in the development of CA, a biologically inspired model which focuses on the trustee's perspective. This new model addresses a serious unresolved problem in existing trust and reputation models, namely the inability to handle constantly changing behaviors and agents' continuous entry and exit from the system. In previous work, we compared CA to FIRE, a well-known trust and reputation model, and found that CA is superior when the trustor population changes, whereas FIRE is more resilient to the trustee population changes. Thus, in this paper, we investigate how the trustors can detect the presence of several dynamic factors in their environment and then decide which trust model to employ in order to maximize utility. We frame this problem as a machine learning problem in a partially observable environment, where the presence of several dynamic factors is not known to the trustor and we describe how an adaptable trustor can rely on a few measurable features so as to assess the current state of the environment and then use Deep Q Learning (DQN), in a single-agent Reinforcement Learning setting, to learn how to adapt to a changing environment. We ran a series of simulation experiments to compare the performance of the adaptable trustor with the performance of trustors using only one model (FIRE or CA) and we show that an adaptable agent is indeed capable of learning when to use each model and, thus, perform consistently in dynamic environments.

## 1. Introduction

In open environments, identifying trustworthy partners to interact with is a challenging task. This issue is usually addressed by trust and reputation mechanisms, which are key elements for the design of Multi Agent Systems (MAS) [3].

However, available trust management methods still have serious weaknesses, such as the inability to deal with agents' mobility and unstable behavior [1]. Common trust models struggle to cope with agents' frequent, unexpected entries and exits [4]. In highly dynamic environments, agents' behavior can change rapidly, and agents using trust mechanisms must be able to quickly detect these changes, in order to select beneficial partners for their interactions [7].

To address these unresolved issues in existing trust and reputation models, we have previously proposed CA [8], a decentralized computational trust model for open MAS, inspired by biological processes in the human brain. Unlike conventional models, CA handles trust from the trustee's point of view. CA draws its strength from letting a trustee decide if it is skilled to provide a service, as required, instead of trustors selecting trustees. In previous work [9], we compared CA to FIRE, an established trust and reputation model, and our main finding was that CA outperforms FIRE when the consumer population changes, while FIRE is more resilient when the trustee population is volatile. Note here, that the terms providers and trustees refer to the agents who provide services, while the terms consumers and trustors refer to the agents who use these services.

This paper focuses on answering the question how trustors may identify the presence of various factors that define a dynamically changing, open MAS (conceptualizing the nature of open MAS appears in [9]), and then decide which trust model to use to maximize utility. This problem can be framed as a machine learning problem in a partially observable environment, in which trustors are unaware of the effect of these dynamically changing factors. This framing renders the problem suitable for a Reinforcement Learning (RL) approach, where an agent learns how to behave in a given environment to maximize rewards. Unlike supervised and unsupervised learning, which use a specific data set to learn from, in RL, agents learn from the rewards and penalties they receive for their actions. Indeed, in our problem, the trustor attempts to learn the optimal policy, i.e., whether to advertise tasks and let trustees decide who gets to carry out a task (as implemented by CA, which is a *push* model) or to select trustees directly (as implemented by FIRE, a *pull* model), with actions being chosen in each state and with utility gain (UG) serving as a reward. We describe in detail how the adaptable trustor can calculate values for a number of environmental variables (features) to assess current state, and use Deep Q Learning (DQN) to learn how to adapt to a changing environment. We ran a series of simulations to compare the performance of an adaptable trustor (using DQN) with the performance of consumers using solely one model, FIRE or CA, and we found that by using Deep Q Learning, the adaptable trustor is able to learn when to use each model, demonstrating a consistently robust performance.

The rest of the paper is organized as follows. In section 2, we review the background on the relevant trust mechanisms and reinforcement learning. Section 3 describes the testbed, DQN architecture, hyper-parameters setup, and the features used for DQN. In section 4, we describe the methodology we employed for our experiments. Section 5 presents our results and section 6 provides a brief discussion of our findings. Finally, in section 7 we conclude our work, highlighting potential future work.

**2. Background**

### 2.1. Trust mechanisms

In this section, we briefly describe the two trust mechanisms used in our simulations: FIRE and CA.

#### 2.1.1. FIRE model

FIRE [5] is an established trust and reputation model, which follows the distributed, decentralized approach, and utilizes the following four sources of trust information:

- *Interaction Trust (IT)*: a target trustee's trustworthiness is evaluated based on the trustor's previous interactions with the target agent.
- *Witness Reputation (WR)*: trustworthiness is estimated by the evaluator based on the opinions of other trustors (witnesses) that have previously interacted with the target trustee.
- *Role-based trust (RT)*: trustworthiness is assessed based on roles and available domain knowledge, including norms and regulations.
- *Certified Reputation (CR)*: trustworthiness is evaluated based on third-party references stored by the target trustee, available on demand.

IT is the most reliable source of trust information because it reflects the evaluator's satisfaction. However, in case that the evaluator has no previous interactions with the target agent, FIRE cannot utilize IT module, and relies on the other three modules, mainly on WR. Nevertheless, in conditions of high flows of witnesses out of the system, WR cannot work properly and FIRE is based mainly on CR module for trust assessments. Yet, CR is not a very reliable source of trust information, since the trustees may choose to store only the best third-party references, resulting in overestimation of their performance.

#### 2.1.2. CA model

We have formally described CA model in [8], as a new computational trust model, inspired by synaptic plasticity, a biological process in human brain. Synaptic plasticity is responsible for the creation of coherent groups of neurons called assemblies (which is why we call our model "Create Assemblies").

CA views trust from the perspective of the trustee, i.e. the trustee decides whether it has the skills to successfully execute a required task. This is opposed to the conventional trust modeling approach, in which the trustor, after gathering and processing behavioral information about possible trustees, finally selects the most reliable one to interact with. As we have previously discussed [8], the idea that the trustor does not select a trustee, provides several advantages to CA approach in open multi-agent systems. The fact that the trustee can carry trust information (stored in the form of connection weights) and use it in every new application it joins, gives CA an advantage in coping with the issue of mobility, which continuous to be recognized as an open challenge [6]. Choosing trustees by trustors increases communication time, because of the apparent need for extensive trust information exchange [12]. Agents may be unwilling to share trust information [11], while revealing an agent's private opinion about others' services may have a detrimental effect

[5]. Finally, in CA approach, agents do not share trust information. This creates the expectation that CA is invulnerable to various types of disinformation, an ongoing problem in most agent societies.

According to CA model, the trustor (the agent requesting the task) broadcasts a request message to all nearby trustees, including the following information: a) the task category (i.e. the type of work to be done), and b) a list of task requirements. When a trustee receives a request message, it creates a connection with a weight $w \in [0,1)$, which represents the connection's strength, i.e. the trust value expressing the trustee's likelihood of successfully performing the task.

After completing the task, the trustee adjusts the weight. If it successfully completes the task, it increases the weight according to equation (1). If the trustee fails, it decreases the weight according to equation (2).

$$w = \text{Min}(1, w + \alpha(1-w)) \quad (1)$$

$$w = \text{Max}(0, w - \beta(1-w)) \quad (2)$$

Positive factors α, β control the rate of increase and decrease, respectively. The trustee takes the decision to execute a task by comparing the connection's weight to a predetermined Threshold $\in [0,1]$; it executes the task only when the weight is not less than the Threshold value.

For our simulation experiments we used CA algorithm for dynamic trustee profiles, as described in [9].

### 2.2. Reinforcement learning

In this setting, the agent interacts with the environment by taking actions and observing their effect (by for example, measuring some environmental quantities). Single-agent RL under full observability is formalized by Sutton & Barto [13], defined as tuple $\langle S, A, T, R, \gamma \rangle$. At timestep t, the agent observes current state $s \in S$, chooses action $a \in A$ based on policy $\pi(a|s)$, and receives reward $r_t = R(s) \in \mathbb{R}$. Then, with probability $P(s'|s,a) = T(s,a,s')$, the environment transits to a new state $s' \in S$. The agent's goal is to learn optimal policy $\pi^*$ maximizing discounted return (future rewards) $R_t = \sum_{t'=t}^{H} \gamma^{t'-t} r_t$, where $H$ is the horizon and $\gamma \in [0,1)$ is a discount factor. The action-value function $Q$ of a policy $\pi$ is defined as $Q^{\pi^*}(s,a) = \mathbb{E}[R_t | S_t = S, a_t = a]$. Then, optimal policy $\pi^*$ maximizes the Q-value function $Q^{\pi^*}(s,a) = \max_\pi Q(s,a)$.

Usually in RL, the agent tries to learn $\pi^*$ without being explicitly given the MDP model. Model-based methods learn $T$ and $R$ and then use a planner to determine $Q^{\pi^*}$. On the contrary, model-free methods are more efficient in terms of space and computation, because they directly learn Q-values or policies.

Q-learning is a model-free method introduced by Watkins and Dayan [14]. It uses backups $Q(s,a) = Q(s,a) + a[r + \gamma \max_{a'} Q(s',a') - Q(s,a)]$ to iteratively calculate the optimal Q-value function, where $a \in [0,1]$ denotes the learning rate. The term

$r + \gamma \max_{a'} Q(s', a') - Q(s, a)$ is the temporal difference (TD) error. In the tabular case, convergence to $Q^{\pi^*}$ is guaranteed, provided adequate state/action space, where other methods using function approximators, such as neural networks are more suitable.

Deep Q-learning [10] is a state-of-the-art method that uses Deep Q-Network (DQN) for Q-value approximation. For any experience $<s, a, r, s'>$, DQN uses two separate neural networks: a) a target neural network $\bar{Q}(\cdot,\cdot;\bar{\theta}_t)$ to calculate the target $r_t + \gamma \max_{a'} \bar{Q}(s', a'; \bar{\theta}_t)$, and b) an online neural network $Q(\cdot,\cdot;\theta)$ to calculate the estimation $Q(s, a; \theta_t)$. Then, the loss $[r_t + \gamma \max_{a'} \bar{Q}(s', a'; \bar{\theta}_t) - Q(s, a; \theta_t)]^2$ is calculated. DQN stores the experiences in a replay memory and updates the online network's parameters at random, taking mini-batches of experiences and minimizing loss using stochastic gradient descent. The target network parameters are updated less frequently to match the online network parameters, allowing a more stable learning.

### 2.3. Towards an adaptable trust mechanism

Can trustor-based and trustee-based schemes co-exist? As stated briefly in the introduction, we can put the burden on trustors to try to detect whether the environment they operate in is one that favors a trustor-based or a trustee-based approach and, based upon that detection, decide what it's the approach they will adopt with the objective to maximize a measure of utility. Of course, as trustors do not necessarily communicate with each other and, when they do so, cannot be guaranteed to communicate in good faith, the question of how to select one's action bears a remarkable affinity to a reinforcement learning context, and, particularly, one in a partially observable environment. Of course, to do so, one has to decide which measurable features will be utilized to describe the environment (more accurately, the allowable state space) and, additionally, how one can use these features to formulate a reward (or, penalty) scheme, that will render the problem solvable by a reinforcement learning technique.

### 3. Experimental setup

This section describes the setup for our simulations. First, in section 3.1, we describe the testbed we designed to meet the needs of our experiments. The DQN architecture and hyper-parameters setup are specified in section 3.2, followed by a description of the features used for state representation in section 3.3.

### 3.1. The testbed

For our simulations, we implemented a testbed similar to the one described in [5]. This section describes its main features and specifications.

The testbed's environment contains agents who deliver services (referred to as trustees or providers) and agents who use these services (referred to as trustors or consumers). For simplicity, all providers offer the same service. The agents are randomly distributed in a spherical world with a radius of 1.0. Radius of operation ($r_0$) indicates the agent's capacity for interactions with other agents, and each agent has other agents situated within its radius of operation, referred to as acquaintances.

A provider's performance varies, determining the utility gain (UG) that a consumer gets from each interaction, calculated as follows. There are four types of providers: bad, ordinary, intermittent, and good. Apart from the intermittent, each provider's actual performance is normally distributed around a mean level of performance $\mu_P$. The values of $\mu_P$ and the related standard deviation $\sigma_P$ for each provider type are shown in Table 1. Intermittent provider has a random performance within the range [PL_BAD, PL_GOOD]. Providers' radius of operation also corresponds to the normal operational range within which they can provide a service without quality loss. In case a consumer is outside that range, the provided service quality decreases linearly in proportion to the distance between the consumer and the provider, but the final calculated performance is always within [-10, +10], and equal to the utility the consumer acquired as an interaction result.

Simulations run in rounds. As in real world, a consumer does not need the service in every round. The probability that the consumer will require the service (activity level $a$) is determined randomly at the agent's creation time. There is no other factor restricting the number of agents eligible to participate in a round. A consumer always requires the service in the round it needs it. The round number is also used as the time value for any event.

There are three consumer groups: a) consumers using only FIRE trust model, b) consumers using only CA, and c) adaptable consumers able to use both models. If a consumer needs the service, first it finds all available nearby providers. FIRE consumers select a provider following the four-step process outlined in [5]. After selecting a provider, FIRE consumers use the service, gain some utility, and rate the service with a rating value equal to the received UG. The rating is then recorded by the consumer for future trust evaluations and the provider is also informed about the rating, which may be stored as evidence of its performance available on demand.

On the contrary, CA consumers that need the service, do not select a provider, but they broadcast a request message, specifying the required service quality. Table 2 presents the five performance levels reflecting the possible qualities of the service. A CA consumer first broadcasts a message requesting the service at the best quality (PERFECT). After sufficient amount of time (WT), all CA consumers, which have not been provided with the service, broadcast a new message requesting the service at the next lower performance level (GOOD). This procedure goes on as long as there exist consumers who have not been served and the requested performance remains above the lowest level. On receiving a request message, a provider saves it locally in a list and runs CA algorithm. WT is a testbed parameter indicating the maximum time needed for all requested services of one round to be delivered.

Adaptable consumer agents first calculate the nine features for the state representation and then use DQN algorithm to decide the trust model (FIRE or CA) that they will use in each simulation round.

In open MAS agents enter and leave the system at any time. This is simulated by replacing a random number of agents, at the end of each round. This number varies, but it is not allowed to exceed a given percentage of the total population. $p_{CPC}$ denotes the consumer population change limit, and $p_{PPC}$ denotes the provider population change limit. The newcomer agents' characteristics are randomly chosen, but the proportions of different consumer groups and provider profiles are maintained.

Altering an agent's location has an effect on both its own situation and its relationship with other agents. Polar coordinates $(r, \varphi, \theta)$ specify an agent's location in the spherical world. To change location, amounts of angular changes $\Delta\varphi$ and $\Delta\theta$ are added to $\varphi$ and $\theta$, respectively. $\Delta\varphi$ and $\Delta\theta$ are chosen at random in $[-\Delta\phi, +\Delta\phi]$. Providers and consumers change locations at the end of a round, with probabilities denoted by $p_{PLC}$ and $p_{CLC}$, respectively.

At the end of each round, the performance $\mu$ of a provider can be changed by an amount $\Delta\mu$ randomly chosen in $[-M, +M]$, with a probability of $p_{\mu C}$. A provider may also switch to a different profile with a probability of $p_{ProfileSwitch}$.

**Table 1 Providers' profiles (performance constants defined in Table 2 )**

| Profile | Range of μp | σp |
| --- | --- | --- |
| Good | [PL_GOOD, PL_PERFECT] | 1.0 |
| Ordinary | [PL_OK, PL_GOOD] | 2.0 |
| Bad | [PL_WORST, PL_OK] | 2.0 |

**Table 2 Performance level constants**

| Performance level | Utility gained |
| --- | --- |
| PL_PERFECT | 10 |
| PL_GOOD | 5 |
| PL_OK | 0 |
| PL_BAD | -5 |
| PL_WORST | -10 |

### 3.2. DQN architecture and hyper-parameters setup

We used neural networks consisting of three fully-connected layers. The input to the neural network is a state representation of nine features' values (section 3.3), thus the input layer has nine neurons, one for each feature. The second, hidden layer has six neurons and the third is the output layer with two distinct neurons, one for each of the two possible actions: *push* (use CA trust model) and *pull* (use FIRE trust model). The choice of activation function for the input and hidden layer is sigmoid, while for the output layer we used linear activation function.

The combinatorial space of hyper-parameters is too large for an exhaustive search. Due to the high computational cost, we have not conducted a systematic search. Instead, an informal search was performed. The values of all hyper-parameters are displayed in Table 3.

**Table 3 List of hyper-parameters and their values**

| Parameter | Value |
| --- | --- |
| n (the width of consecutive rounds in features' calculation) | 10 |
| Constant ϵ for exploration | 0.05 |
| λ for L2 regularization | 0.01 |
| Learning rate in DQN algorithm | 0.3 |
| Backward propagation learning rate | 0.15 |
| Discount factor γ | 0.95 |
| Replay memory size | 50 |
| Minibatch size | 5 |
| Steps between target network updates | 5 |

### 3.3. Features

In this section, we describe the nine features we used for state representation, which are summarized in Table 4.

**Table 4 Selected Features for the DQN**

| Feature |
|---|
| ProvidersPopulationDirectChangeEstimate |
| MeanProvidersPopulationDirectChangeEstimate |
| ProvidersPopulationIndirectChangeEstimate |
| MeanProvidersPopulationIndirectChangeEstimate |
| NewcomerEstimate |
| ConsumersPopulationIndirectChangeEstimate |
| MeanConsumersPopulationIndirectChangeEstimate |
| ProvidersPerformanceChangeEstimate |
| ConsumersLocationChangeEstimate |

The trustor's decision on which model is most appropriate at any given time is based on the accuracy of its assessment about the environment's current state, which is determined by the presence of the following factors:

- The provider population change
- The consumer population change
- The provider's average level of performance change
- The provider's change into a different performance profile
- The provider's move to a new location on the spherical world
- The consumer's move to a new location on the spherical world

Since we assume partially observable environment, meaning that the trustor has no knowledge of the aforementioned factors, the trustor can only utilize the following available data:

- A local rating database. After each interaction, the trustor rates the provided service and each rating is stored in its local rating database.
- The trustor's acquaintances: the agents situated in the trustor's radius of operation.
- The trustor's nearby provider agents: the providers situated in the trustor's radius of operation.

- The trustor is able to evaluate the trustworthiness of all nearby providers using the FIRE model. Providers whose trustworthiness cannot be determined (for any reason) are placed in the set NoTrustValue. The rest, whose trustworthiness can be determined are placed in the set HasTrustValue.
- The trustor's location on the spherical world, i.e. its polar coordinates $(r, \varphi, \theta)$.
- The acquaintances' ratings or referrals, according to the process of Witness Reputation (WR): when consumer agent a evaluates agent $b$'s WR, it sends a request for ratings to $n_{BF}$ acquaintances, that are likely to have ratings for agent b. Upon receiving the query, each acquaintance will attempt to match it to its own ratings database and return any relative ratings. In case acquaintance cannot find the requested ratings (because it has had no interactions with b), it will only return referrals identifying its $n_{BF}$ acquaintances.
- Nearby providers' certified ratings, according to the process of Certified Reputation (CR): After each interaction, provider $b$ requests from its partner consumer $c$ to provide certified ratings for its performance. Then, $b$ selects the best ratings to be saved in its local rating database. When $a$ expresses its interest for $b$'s services, it asks $b$ to give references about its previous performance. Then, agent $a$ receives $b$'s certified rating set and uses it to calculate $b$'s CR.

Next, we elaborate on how the trustor can approximate the real environment's conditions. We enumerate indicative changes in the environment and analyze on how the trustor could sense this particular change and measure the extent of the change.

1. *The provider population changes at maximum X% in each round*.

Given that $nearbyProviders_t$ is the list of nearby providers in round $t$, and $nearbyProviders_{t-1}$ is the list of nearby providers in round $t-1$, the consumer can calculate the list of the newcomer nearby provider agents $newProviders_t$, as the providers that exist in $nearbyProviders_t$ ($newProviders_t \in nearbyProviders_t$), but not in $nearbyProviders_{t-1}$ ($newProviders_t \notin nearbyProviders_{t-1}$). Then, the following feature can be calculated by the consumer agent as an index of the provider population change in each round:

$$\text{ProvidersPopulationDirectChangeEstimate}_t = \frac{|newProviders_t|}{|nearbyProviders_t|} \quad (3)$$

where the meaning of $|\cdot|$ is cardinality of. Intuitively, this ratio expresses how changes in the total provider population can be reflected in changes in the provider population situated in the consumer's radius of operation. However, in any given round, newcomer providers may not be located within a specific consumer's radius of operation, even if new providers have entered the system. In this case, the total change of the providers' population in the system can be better assessed in a window of consecutive rounds.

Given that a window *of rounds* $W$ is defined as a set of $n$ consecutive rounds $W = \{t_1, t_2, \cdots, t_n\}$, the consumer can calculate feature $ProvidersPopulationDirectChangeEstimate$ for each round $t_i \in W$, as follows.

$$\text{ProvidersPopulationDirectChangeEstimate}_{t_i} = \frac{\left|newProviders_{t_i}\right|}{\left|nearbyProviders_{t_i}\right|} \quad (4)$$

Then, a more accurate index of the provider population change in each round can be calculated as the following mean:

$$\text{MeanProvidersPopulationDirectChangeEstimate}_{t_n} = \frac{\sum_{i=1}^{n} ProvidersPopulationDirectChangeEstimate_{t_i}}{n} \quad (5)$$

Note that $newProviders_t$ can include providers that are not really newcomers, but they previously existed in the system and appear as new agents in $nearbyProviders_t$, either due to their own movement in the world, or due to the consumer's movement, or both. In other words, agents' movement in the world introduce noise in the measurement of $ProvidersPopulationDirectChangeEstimate$.

Each newcomer provider agent has not yet interacted with other agents in the system and therefore its trustworthiness cannot be determined using Interaction Trust and Witness Reputation modules. If the newcomer provider does not have any certified ratings from its interactions with agents in other systems, then its trustworthiness cannot be determined using Certified Reputation module, either. This agent belongs to the $NoTrustValue_{t_i}$ set, which is a subset of $nearbyProviders_{t_i}$.

We can distinguish another category of newcomer provider agents, those whose trustworthiness can be determined solely by CR, because they own certified ratings from their interactions with agents in other systems. Let $A_{t_i}$ denote the subset of $nearbyProviders_{t_i}$ whose trustworthiness can be determined using only the CR module. The consumer can then calculate the following feature as an alternative index for the provider population change in a round.

$$\text{ProvidersPopulationIndirectChangeEstimate}_{t_i} = \frac{\left|NoTrustValue_{t_i}\right| + \left|A_{t_i}\right|}{\left|nearbyProviders_{t_i}\right|} \quad (6)$$

As previously stated, in any given round, newcomer providers may not be located within a specific consumer's radius of operation, even if they have entered the system. Thus, a more accurate estimate of ProvidersPopulationIndirectChangeEstimate can be calculated in a window of n consecutive rounds, with the following mean:

$$\text{MeanProvidersPopulationIndirectChangeEstimate}_{t_n} = \frac{\sum_{i=1}^{n} ProvidersPopulationIndirectChangeEstimate_{t_i}}{n} \quad (7)$$

2. *The consumer population changes at maximum X% in each round*.

If the consumer population has changed, it is possible that the consumer itself is a newcomer. In this case, the consumer is not expected to have ratings for any of its nearby providers in its local rating database, and IT module of FIRE cannot work. Thus, estimating whether it is a newcomer consumer probably helps the agent to decide about which trust model to employ for maximum benefits. The consumer can calculate feature NewcomerEstimate as follows. It sets $NewcomerEstimate = 1$, if there are no available ratings in its local rating database for any of the nearby providers, otherwise it sets $NewcomerEstimate = 0$. More formally:

$$NewcomerEstimate = \begin{cases} 1, & if\ \forall\ provider \in nearbyProviders: \nexists\ rating = (consumer, provider, \_, \_, \_) \in ratingDatabase_{consumer} \\ 0, & otherwise \end{cases} \quad (8)$$

According to WR process, when consumer $a$ assesses the WR of a nearby provider $b$, it sends out a query for ratings to $n_{BF}$ consumer acquaintances that are likely to have relative ratings on agent b. Each newcomer consumer acquaintance will try to match the query to its own (local) rating database, but it will find no matching ratings, because of no previous interactions with b. Thus, the acquaintance will return no ratings. In this case, the consumer can reasonably infer that an acquaintance that returns no ratings is very likely to be a newcomer agent. So, consumer agent a, in order to approximate the consumer population change, can calculate the following feature:

$$\text{ConsumersPopulationIndirectChangeEstimate}_{t_i} = \frac{|B_{t_i}|}{|C_{t_i}|} \quad (9), \text{ where:}$$

$C_{t_i}$ is the set of consumer acquaintances of $a$ to which consumer $a$ sends queries (for ratings), one for each of its nearby providers, at round $t_i$, and $B_{t_i}$ is the subset of $C$ consumer acquaintances that returned no ratings (for any query) to $a$ at round $t_i$.

As previously stated, in any given round, newcomer consumer acquaintances may not be located within a specific consumer's radius of operation, even if newcomer consumers have entered the system. Thus, a more accurate estimate of ConsumersPopulationIndirectChangeEstimate can be calculated in a window of $n$ consecutive rounds, with the following mean:

$$\text{MeanConsumersPopulationIndirectChangeEstimate}_{t_n} = \frac{\sum_{i=1}^{n} ConsumersPopulationIndirectChangeEstimate_{t_i}}{n} \quad (10)$$

3. *The provider alters its average level of performance at maximum X UG units with a probability of p in each round*.

A consumer using FIRE, selects the provider with the highest trust value in the HasTrustValue set. The quantity $trustValue * 10$ is an estimate of the average performance of the selected provider whose actual performance can be different. The difference $|trustValue * 10 - actualPerformance|$ expresses the change of the average performance level of a specific provider. Yet, not all providers alter their average level of performance in each round. Thus, the consumer can calculate the following feature as an estimate of a provider's average level of performance, by using the differences $|trustValue_i * 10 - actualPerformance_i|$ of the last N interactions with this provider.

$$\text{ProvidersPerformanceChangeEstimate} = \frac{\sum_{i}^{N} |trustValue_i * 10 - actualPerformance_i|}{N} \quad (11)$$

4. *The providers switch into a different (performance) profile with a probability of p in each round*

A consumer can use ProvidersPerformanceChangeEstimate as an estimate of a provider's profile change, as well.

5. *Consumers move to a new location on the spherical world with a probability p in each round.*

The list $nearbyProviders$ can include providers that appear as new because of the consumer's movement on the world. Thus, $ProvidersPopulationDirectChangeEstimate_{t_i}$ can be also be used as an estimate of consumer's movement.

Nevertheless, in order to make a more accurate estimate, a consumer can use its polar coordinates. Given that $(r_t, \varphi_t, \theta_t)$ are its polar coordinates in round $t$ and $(r_{t-1}, \varphi_{t-1}, \theta_{t-1})$ are its polar coordinates in round $t - 1$, a consumer is able to calculate the following feature:

$$\text{ConsumersLocationChangeEstimate} = \begin{cases} 1, if\ \Delta r = |r_t - r_{t-1}| \neq 0 \vee \Delta\varphi = |\varphi_t - \varphi_{t-1}| \neq 0 \vee \Delta\theta = |\theta_t - \theta_{t-1}| \neq 0 \\ 0, otherwise \end{cases} \quad (12)$$

The consumer estimates that its location has changed, by setting $ConsumersLocationChangeEstimate = 1$, if at least one of its polar coordinates $r, \varphi, \theta$ have changed, compared to the previous round.

6. Providers move to a new location on the spherical world with a probability p in each round

The list $nearbyProviders$ can include providers that appear as new because of their movement in the world. Thus, $ProvidersPopulationDirectChangeEstimate_{t_i}$ also serves as an estimate of the providers' movement.

Note that since a provider does not know other agents' polar coordinates, it cannot calculate a feature similar to $ConsumersLocationChangeEstimate$.

4. **Experimental methodology**

In our simulation experiments, we compare the performance of the following three consumer groups:

- Adaptable: consumer agents able to choose a trust model (FIRE or CA) based on their assessment for the presence of several dynamic factors in their environment.
- FIRE: consumer agents using only FIRE trust model.
- CA: consumers using solely CA trust model.

To accomplish this, we run a number of independent simulation runs (NISR) to ensure more accurate results and avoid random noise. In order to obtain statistically significant results, NISR varies in each experiment, as shown in Table 5.

The utility that each agent gained throughout simulations indicates the model's ability to identify reliable, profitable providers. Thus, each interaction's utility gain (UG), along with the trust model employed are both recorded.

After all simulations runs have finished, we calculate the average UG for each interaction for each consumer group. The average UG of the three consumer groups are then compared using the two-sample t-test for means comparison [2] with a 95% confidence level.

The results of each experiment are presented in a separate graph showing the UG means in each interaction, for each consumer group. In all the simulations the provider population is "typical", as defined in [5], consisting of half profitable providers (producing positive UG) and half harmful providers (producing negative UG, including intermittent providers).

The values for the experimental variables used are shown in Table 6, while the FIRE and CA parameters used are shown in Table 7 and Table 8, accordingly.

**Table 5 Number of independent simulation runs (NISR) per experiment**

| Experiment | NISR |
|---|---|
| 1, 2, 6, 8-11, 17 | 30 |
| 3-5, 7, 12, 14-16, 18 | 10 |
| 13 | 12 |

**Table 6 Experimental variables**

| Simulation variable | Symbol | Value |
|---|---|---|
| Number of simulation rounds | $N$ | |
| - Default | | 500 |
| - Experiments: 7, 13 | | 1000 |
| Total number of provider agents: | $N_P$ | 100 |
| Good providers | $N_{GP}$ | 10 |
| Ordinary providers | $N_{PO}$ | 40 |
| Intermittent providers | $N_{PI}$ | 5 |
| Bad providers | $N_{PB}$ | 45 |
| Total number of consumer agents | $N_C$ | 500 |
| Range of consumer activity level | $\alpha$ | [0.25, 1.00] |
| Waiting Time | WT | 1000 msec |

**Table 7 FIRE's default parameters**

| Parameters | Symbol | Value |
|---|---|---|
| Local rating history size | $H$ | 10 |
| IT recency scaling factor | $\lambda$ | -(5/ln(0.5)) |
| Branching factor | $n_{BF}$ | 2 |
| Referral length threshold | $n_{RL}$ | 5 |
| Component coefficients | | |
| Interaction trust | $W_I$ | 2.0 |
| Role-based trust | $W_R$ | 2.0 |
| Witness reputation | $W_W$ | 1.0 |
| Certified reputation | $W_C$ | 0.5 |
| Reliability function parameters | | |
| Interaction trust | $\gamma_I$ | -ln(0.5) |
| Role-based trust | $\gamma_R$ | -ln(0.5) |
| Witness reputation | $\gamma_W$ | -ln(0.5) |
| Certified reputation | $\gamma_C$ | -ln(0.5) |

**Table 8 CA's default parameters**

| Parameters | Symbol | Value |
|---|---|---|
| Threshold | *Threshold* | 0.5 |
| Positive factor controlling the rate of the increase in strengthening of a connection | $\alpha$ | 0.1 |
| Positive factor controlling the rate of the decrease in weakening of a connection | $\beta$ | 0.1 |

## 5. Simulation results
### 5.1. Adaptable's performance in single environmental changes

In this section we compare the performance of three groups of consumers (adaptable, CA, FIRE), when there is no environmental change (static setting) and when there is one environmental change at a time, keeping the change constant throughout simulation, conducting the following experiments:

Experiment 1. The setting is static, with no changes.

Experiment 2. The provider population changes at maximum 2% in every round ($p_{PPC} = 0.02$).

Experiment 3. The provider population changes at maximum 5% in every round ($p_{PPC} = 0.05$).

Experiment 4. The provider population changes at maximum 10% in every round ($p_{PPC} = 0.10$).

Experiment 5. The consumer population changes at maximum 2% in every round ($p_{CPC} = 0.02$).

Experiment 6. The consumer population changes at maximum 5% in every round ($p_{CPC} = 0.05$).

Experiment 7. The consumer population changes at maximum 10% in every round ($p_{CPC} = 0.10$).

Experiment 8. A provider may alter its average level of performance at maximum 1.0 UG unit with a probability of 0.10 each round ($p_{\mu C} = 0.10, M = 1.0$).

Experiment 9. A provider may switch into a different (performance) profile with a probability of 2% in every round ($p_{ProfileSwitch} = 0.02$).

Experiment 10. A consumer may move to a new location on the spherical world at a maximum angular distance of $\pi/20$ with a probability of 0.10 in every round ($p_{CLC} = 0.10, \Delta\Phi = \pi/20$).

Experiment 11. A provider may move to a new location on the spherical world at a maximum angular distance of $\pi/20$ with a probability of 0.10 in every round ($p_{PLC} = 0.10, \Delta\Phi = \pi/20$).

The experiments' results are presented in figures Fig. 1 - Fig. 11. Overall, the performance of the adaptable consumer is satisfactory in all the experiments and follows the performance of the best-choice model in each situation.

This conclusion is drawn more easily, if we consider experiments 3-7 (figures Fig. 3 - Fig. 7), where the performance of the adaptable consumer is always between the performance of the other two models. Specifically, when the provider population changes (figures Fig. 3Fig. 4), the adaptable's performance is always higher than CA, the worst choice in this case. On the contrary, when the consumer population is volatile (figures Fig. 5 - Fig. 7), the adaptable maintains a performance higher than that of FIRE's.

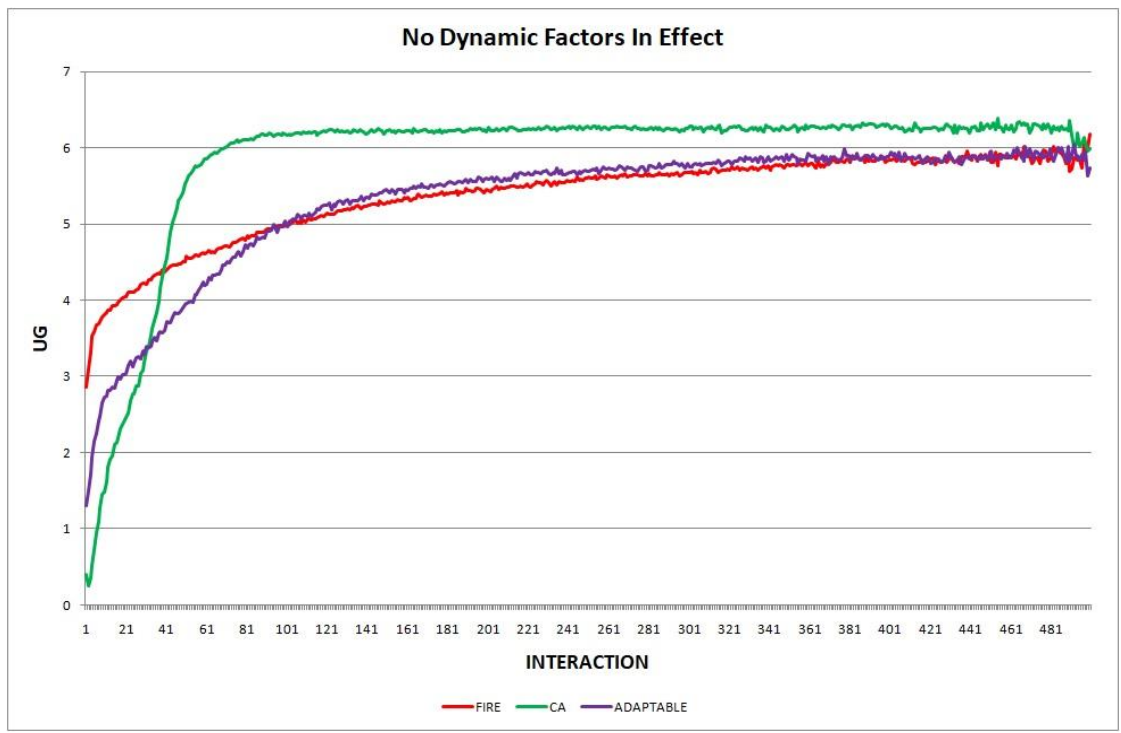

Fig. 1 Experiment 1. Performance of the adaptable consumer in the static setting

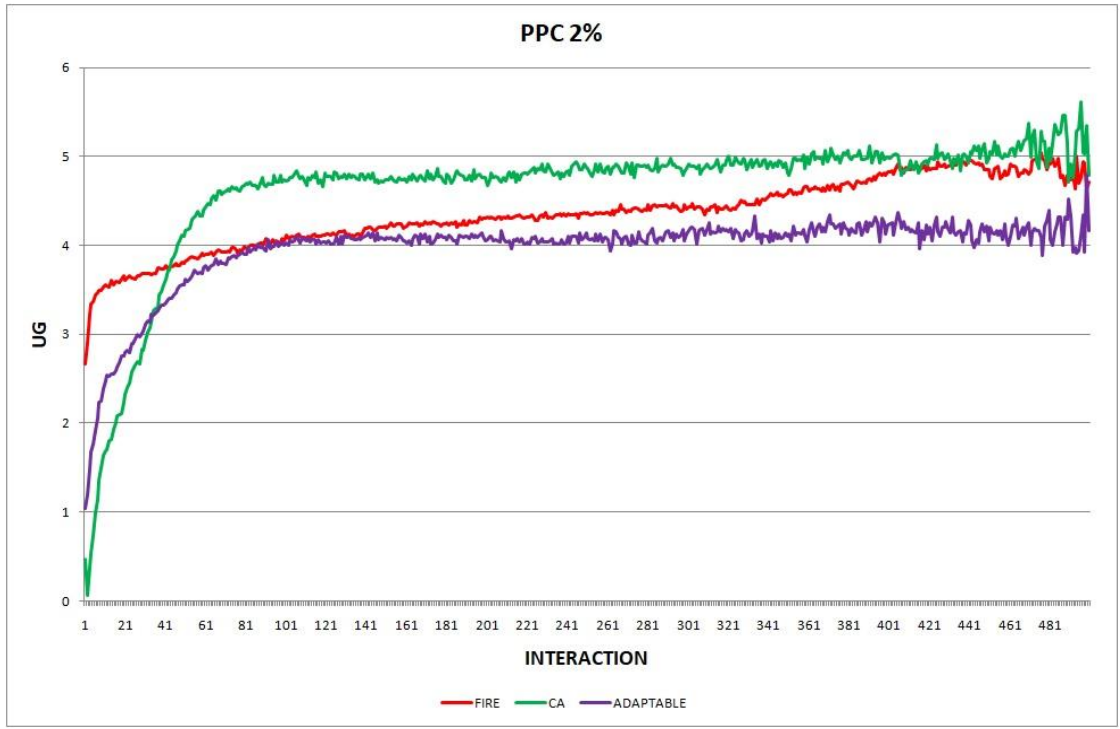

Fig. 2 Experiment 2: Provider population change $p_{PPC} = 2\%$

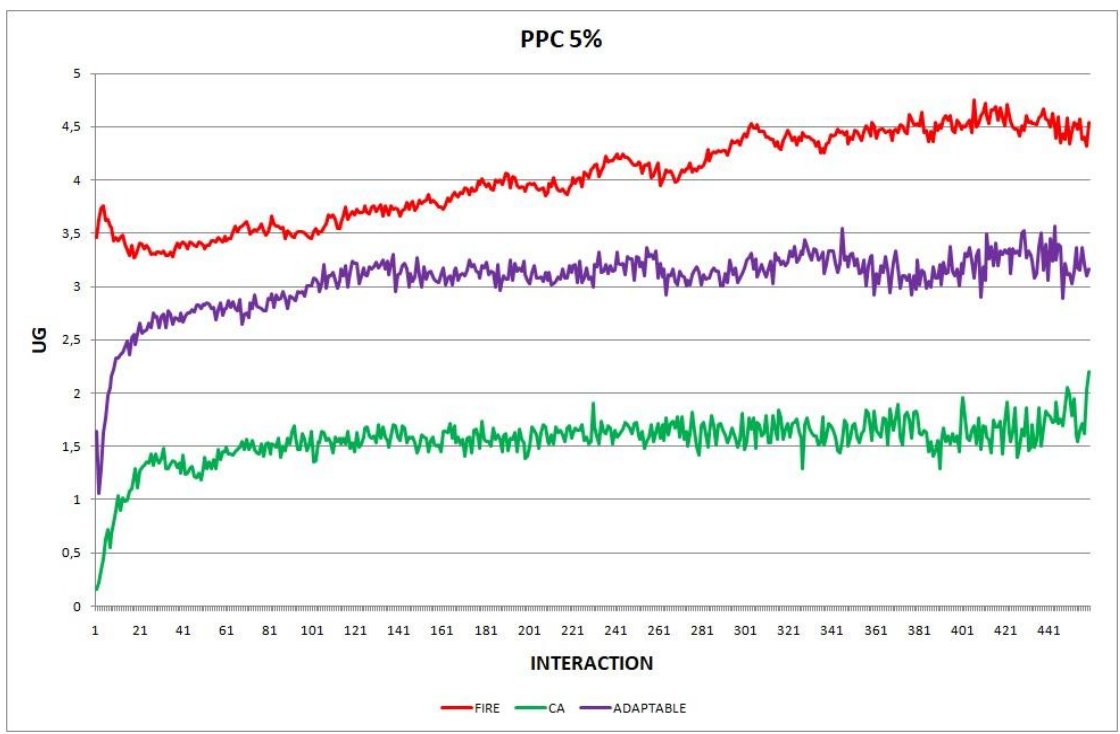

Fig. 3 Experiment 3: Provider population change $p_{PPC} = 5\%$

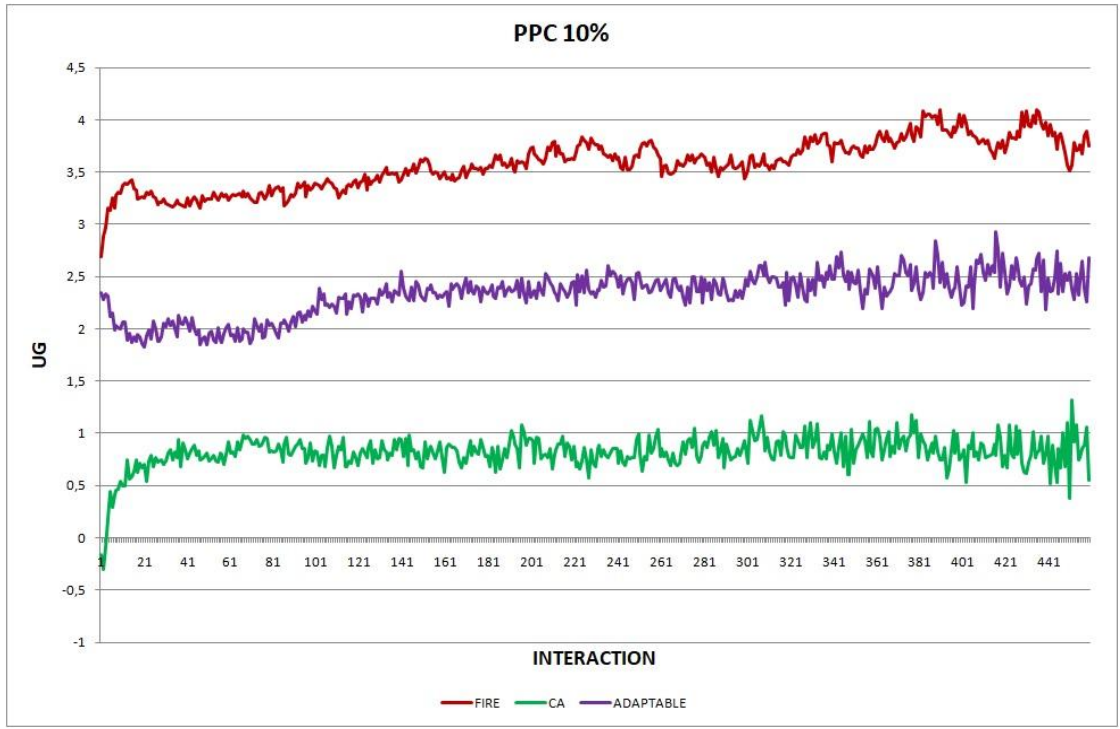

Fig. 4 Experiment 4: Provider population change $p_{PPC} = 10\%$

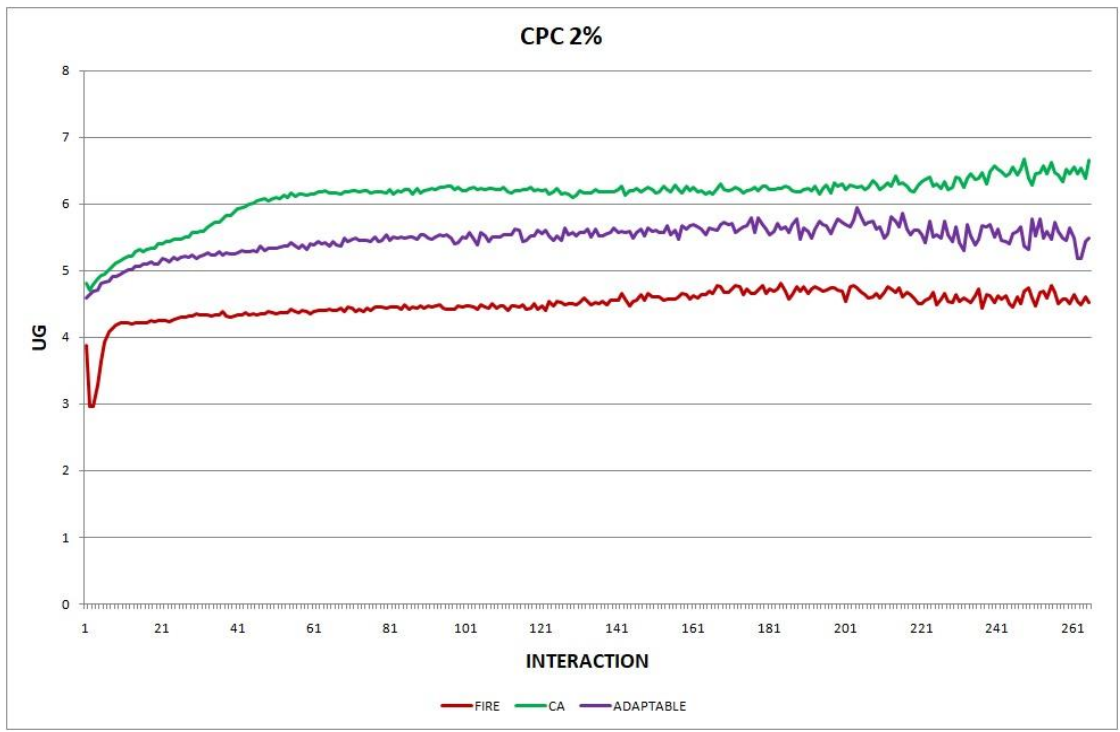

Fig. 5 Experiment 5: Consumer population change $p_{CPC} = 2\%$

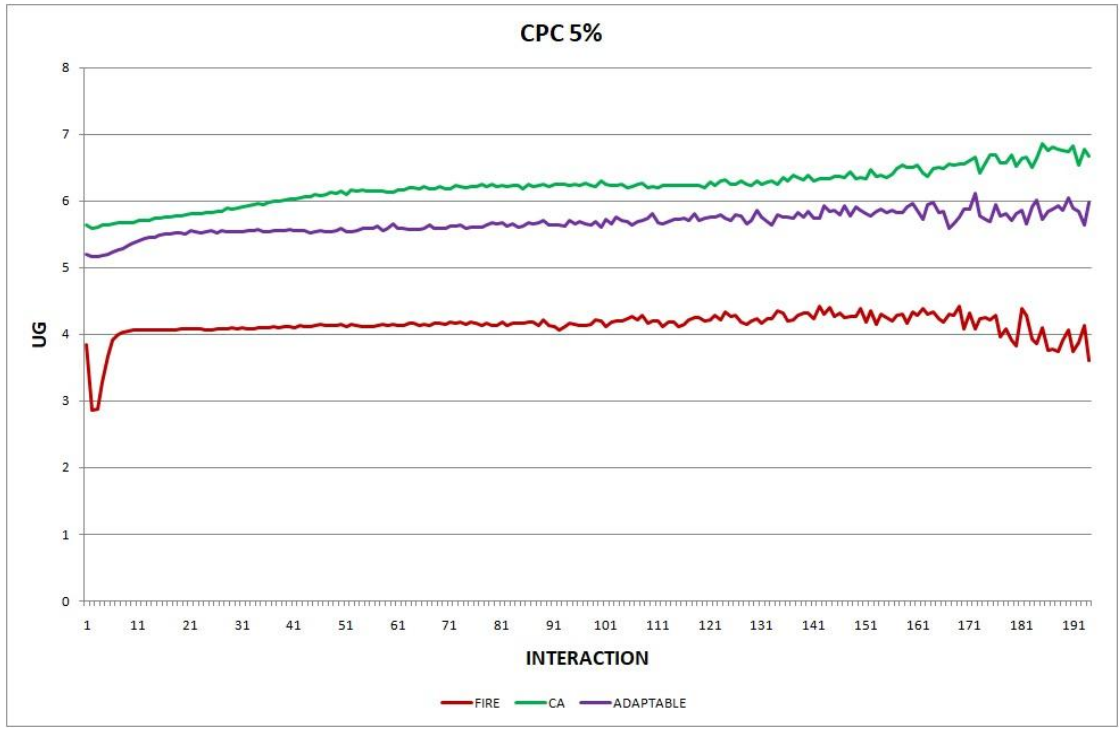

Fig. 6 Experiment 6: Consumer population change $p_{CPC} = 5\%$

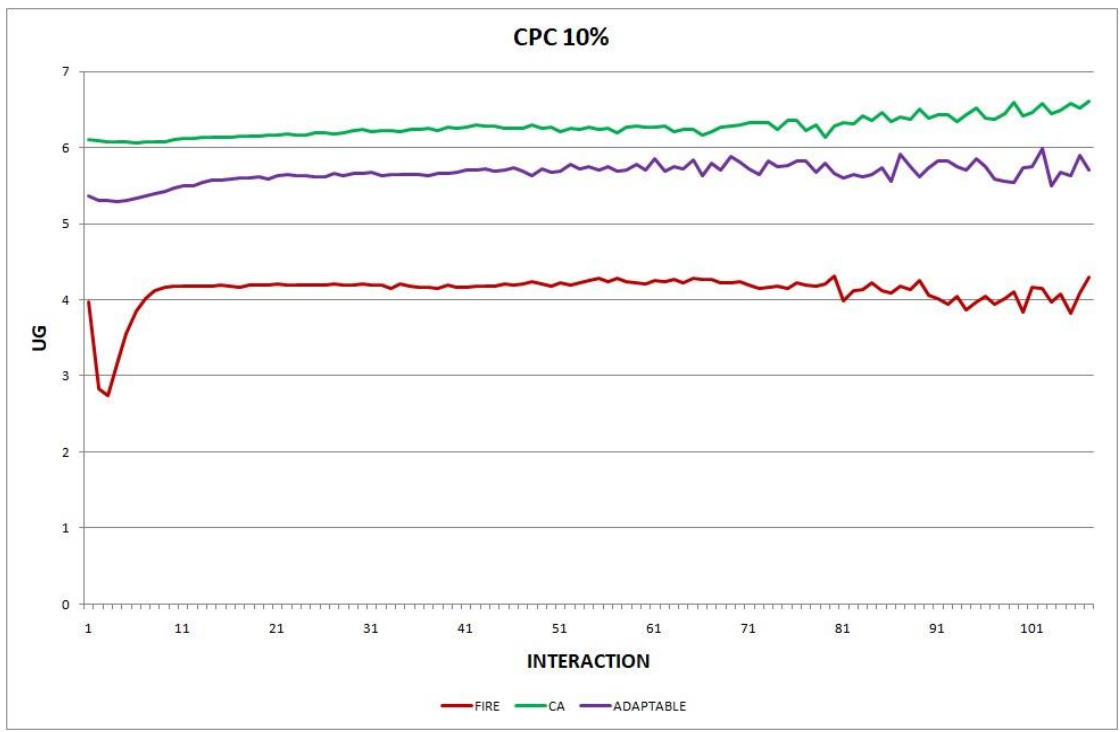

Fig. 7 Experiment 7: Consumer population change $p_{CPC} = 10\%$

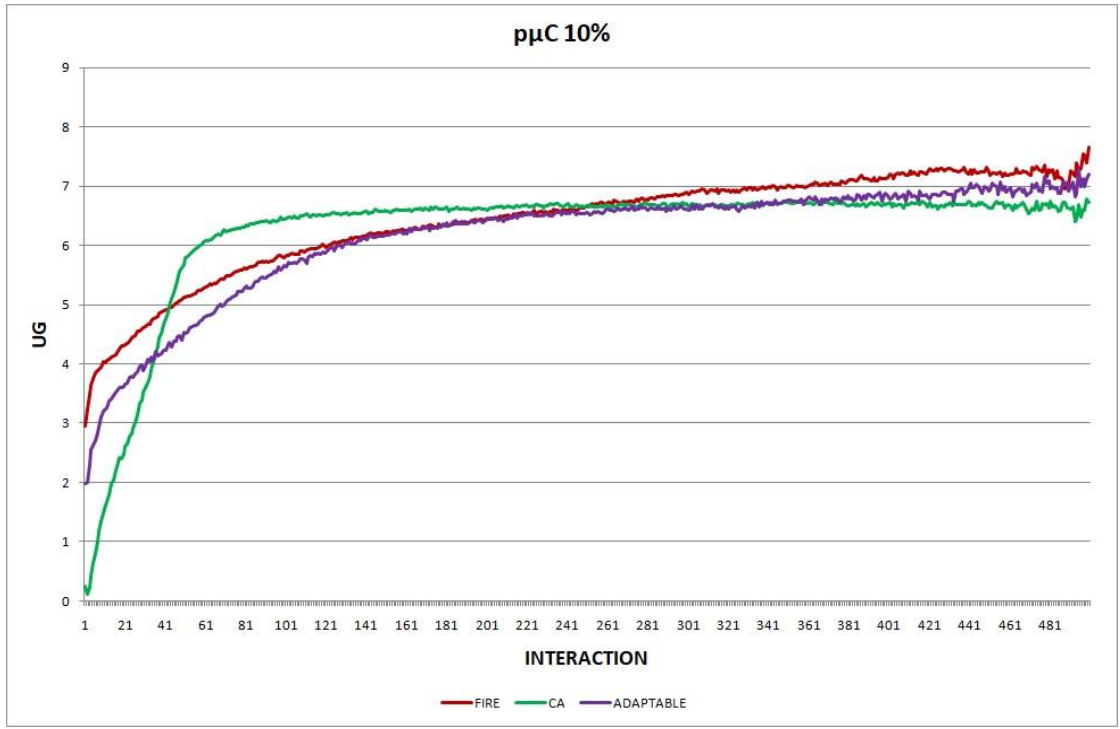

Fig. 8 Experiment 8: Providers change their performance: $p_{\mu C} = 10\%, M = 1.0$

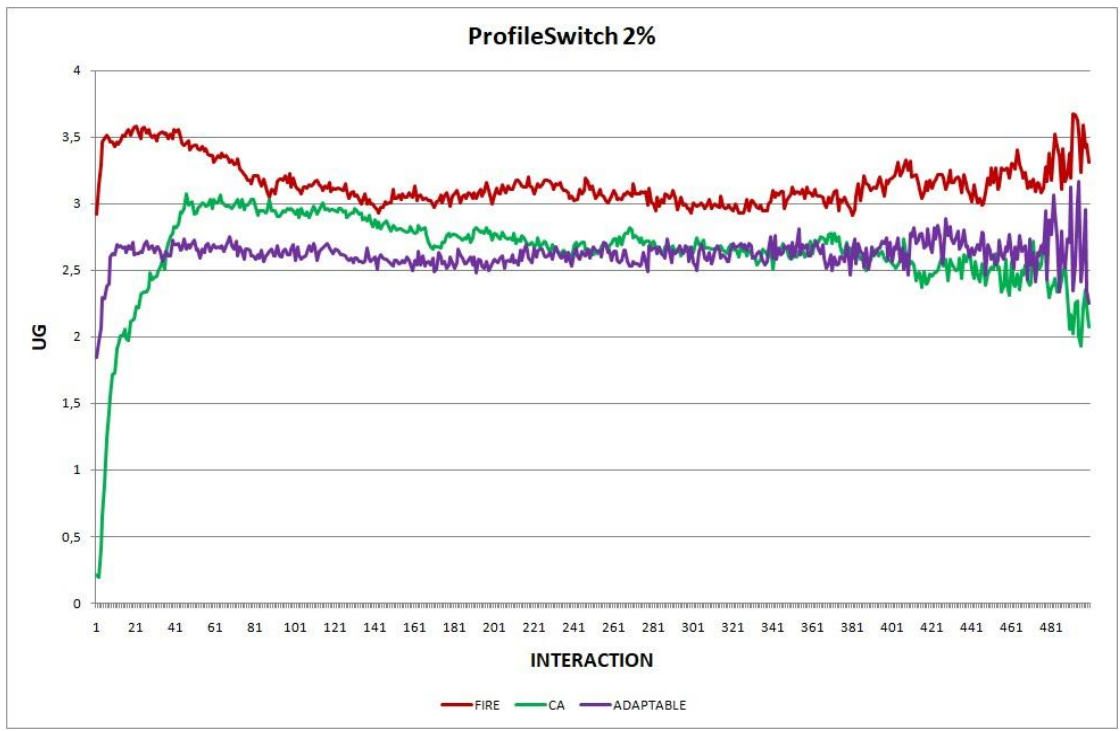

Fig. 9 Experiment 9: Providers switch their profiles: $p_{ProfileSwitch} = 2\%$

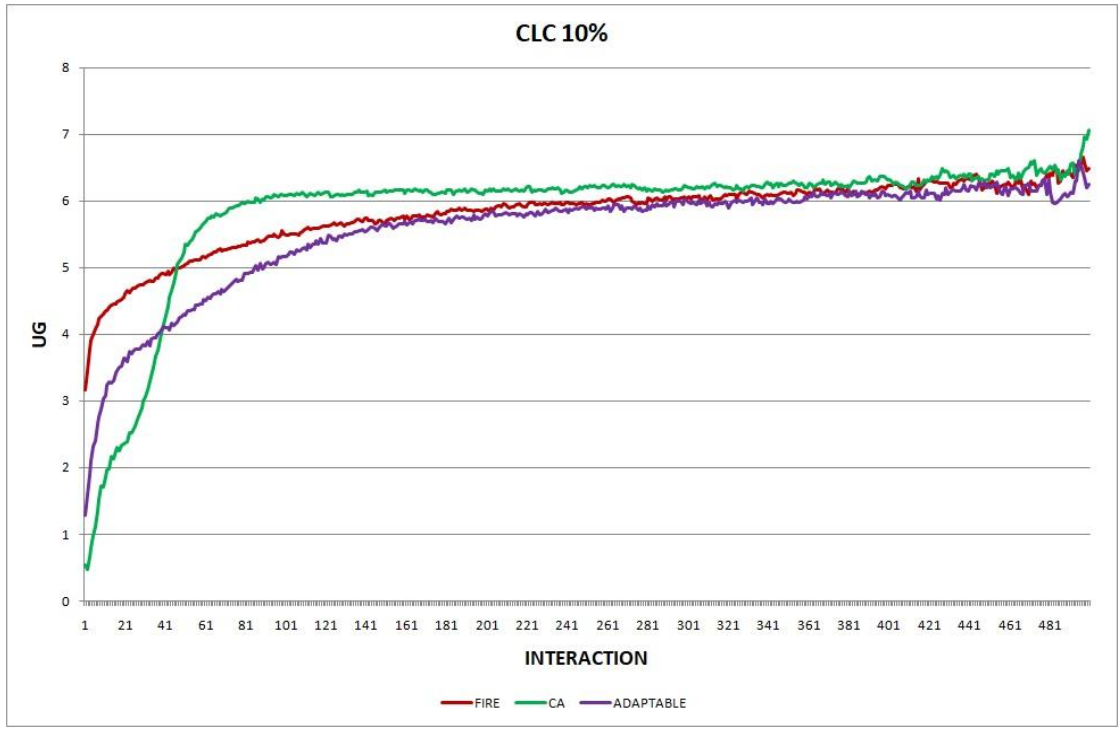

Fig. 10 Experiment 10: Consumers change their locations: $p_{CLC} = 10\%, \Delta\Phi = \pi/20$

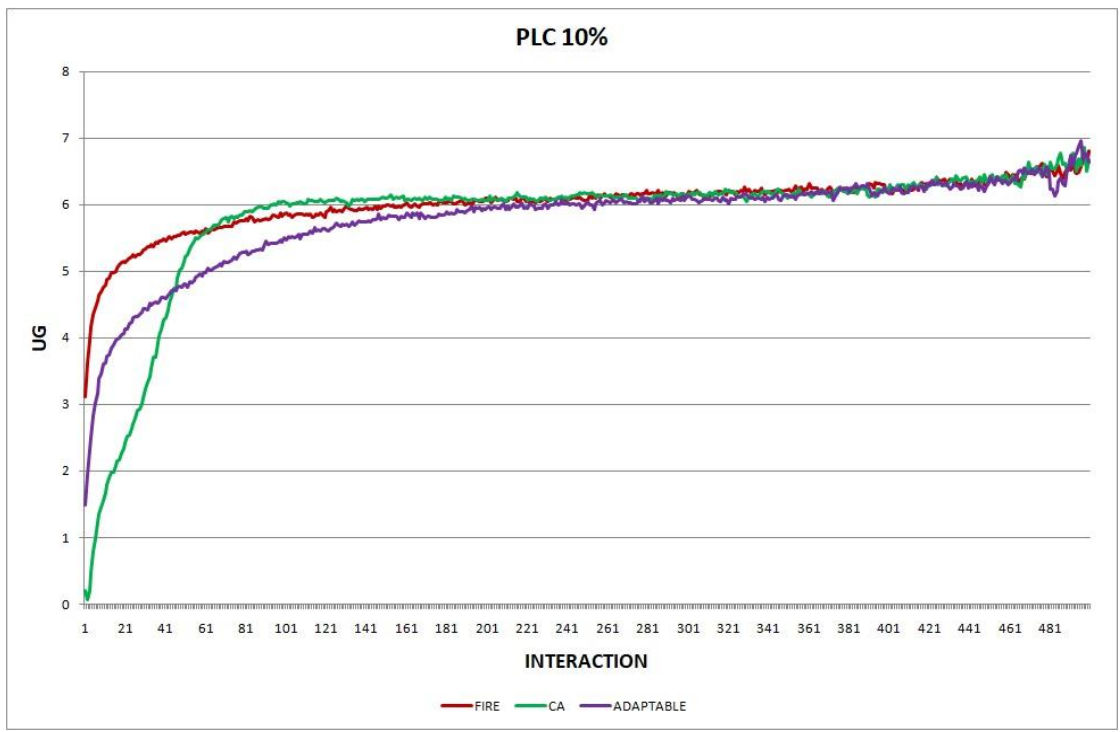

Fig. 11 Experiment 11: Providers change their locations: $p_{PLC} = 10\%, \Delta\Phi = \pi/20$

### 5.2. Adaptable's performance in various combinations of changes

The experiments of this section aim to demonstrate that the adaptable consumer can find the optimal policy when there are multiple, concurrently acting environmental changes and it is difficult to predict (using previous section's results) which of the two models FIRE or CA, is the best option. Note, that all changes are maintained the same throughout simulation rounds of the experiment. The experiments we conducted are described as follows.

Experiment 12. The provider population changes at maximum 2% in every round ($p_{PPC} = 0.02$) and the consumer population changes at maximum 5% in every round ($p_{CPC} = 0.05$).

Experiment 13. The provider population changes at maximum 10% in every round ($p_{PPC} = 0.02$), the consumer population changes at maximum 10% in every round ($p_{CPC} = 0.05$).

Experiment 14. The provider population changes at maximum 2% in every round ($p_{PPC} = 0.02$), the consumer population changes at maximum 5% in every round ($p_{CPC} = 0.05$) and consumers may move to a new location on the spherical world at a maximum angular distance of π/20 with a probability of 0.10 in every round ($p_{CLC} = 0.10, \Delta\Phi = \pi/20$).

Experiment 15. The provider population changes at maximum 2% in every round ($p_{PPC} = 0.02$), the consumer population changes at maximum 5% in every round ($p_{CPC} = 0.05$) and both consumers and providers may move to a new location on the spherical world at a maximum angular distance of π/20 with a probability of 0.10 in every round ($p_{CLC} = 0.10, p_{PLC} = 0.10, \Delta\Phi = \pi/20$).

Experiment 16. The provider population changes at maximum 2% in every round ($p_{PPC} = 0.02$), the consumer population changes at maximum 5% in every round ($p_{CPC} = 0.05$), both consumers and providers may move to a new location on the spherical world at a maximum angular distance of π/20 with a probability of 0.10 in every round ($p_{CLC} = 0.10, p_{PLC} = 0.10, \Delta\Phi = \pi/20$), and providers may alter their average level of performance at maximum 1.0 UG unit with a probability of 0.10 each round ($p_{\mu C} = 0.10, M = 1.0$).

Experiment 17. The provider population changes at maximum 2% in every round ($p_{PPC} = 0.02$), the consumer population changes at maximum 5% in every round ($p_{CPC} = 0.05$), both consumers and providers may move to a new location on the spherical world at a maximum angular distance of π/20 with a probability of 0.10 in every round ($p_{CLC} = 0.10, p_{PLC} = 0.10, \Delta\Phi = \pi/20$), providers may alter their average level of performance at maximum 1.0 UG unit with a probability of 0.10 each round ($p_{\mu C} = 0.10, M = 1.0$) and providers may switch into a different (performance) profile with a probability of 2% in every round ($p_{ProfileSwitch} = 0.02$).

Overall, in no experiment does the performance of the adaptable fall below the performance of the model with the worst performance, which supports the conclusion that the adaptable indeed succeeds in finding the optimal policy in all the experiments. Interestingly, in Experiment 12 (figure Fig. 12) the performance of the adaptable consumer group outperforms the other two consumer groups (FIRE and CA).

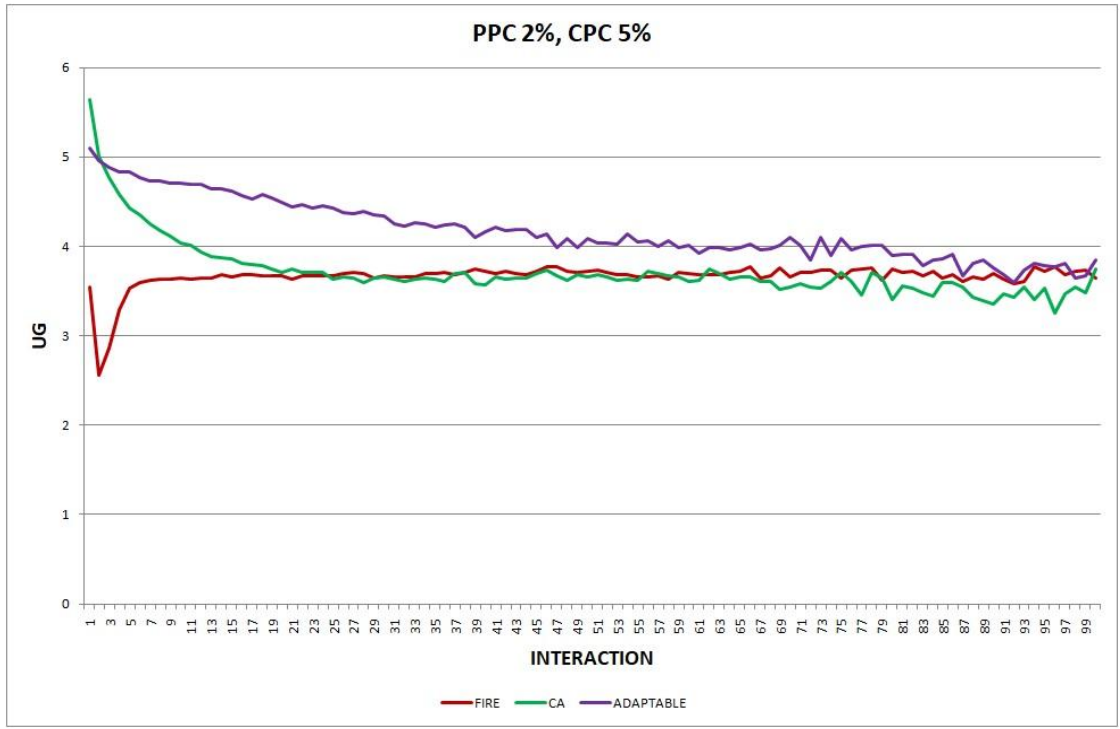

Fig. 12 Experiment 12: $p_{PPC} = 2\%$ and $p_{CPC} = 5\%$.

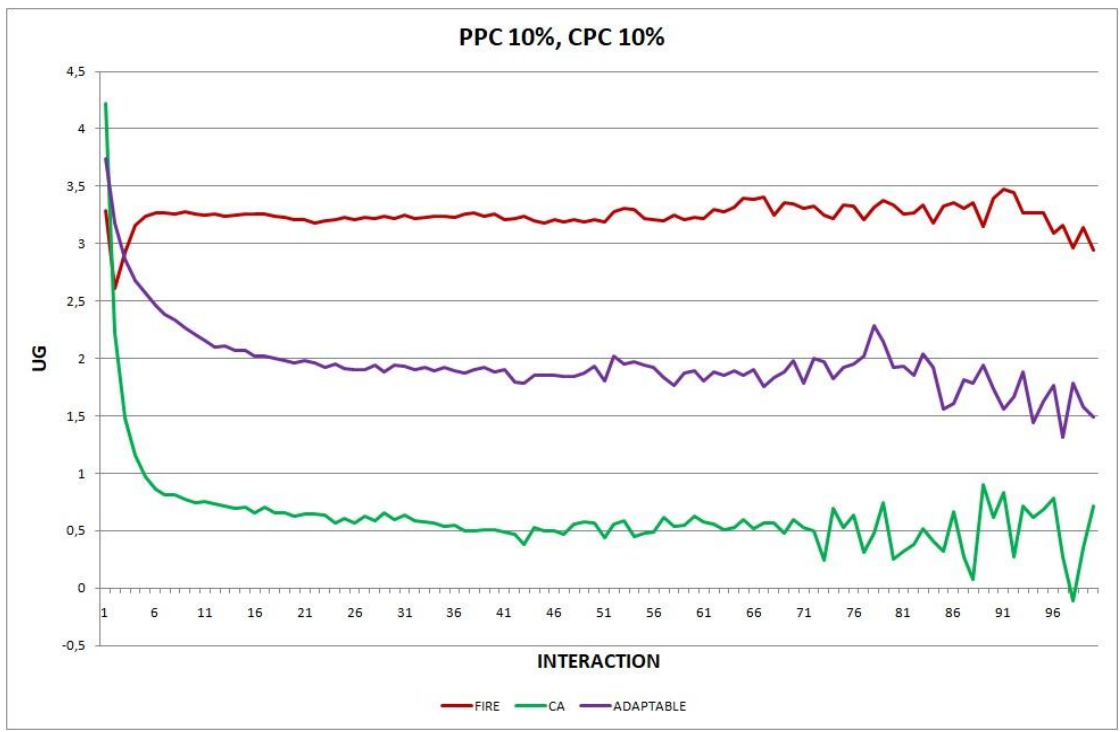

Fig. 13 Experiment 13: : $p_{PPC} = 10\%$ and $p_{CPC} = 10\%$.

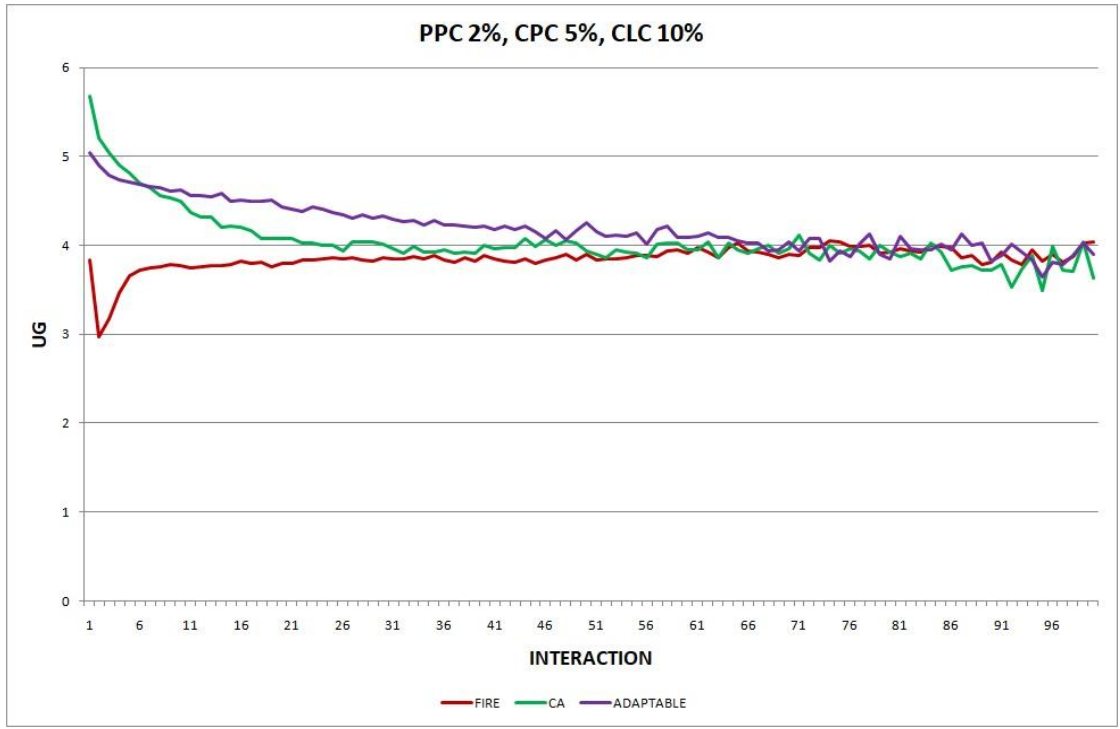

Fig. 14 Experiment 14: $p_{PPC} = 2\%, \ p_{CPC} = 5\%, \ p_{CLC} = 10\%, \Delta\Phi = \pi/20$

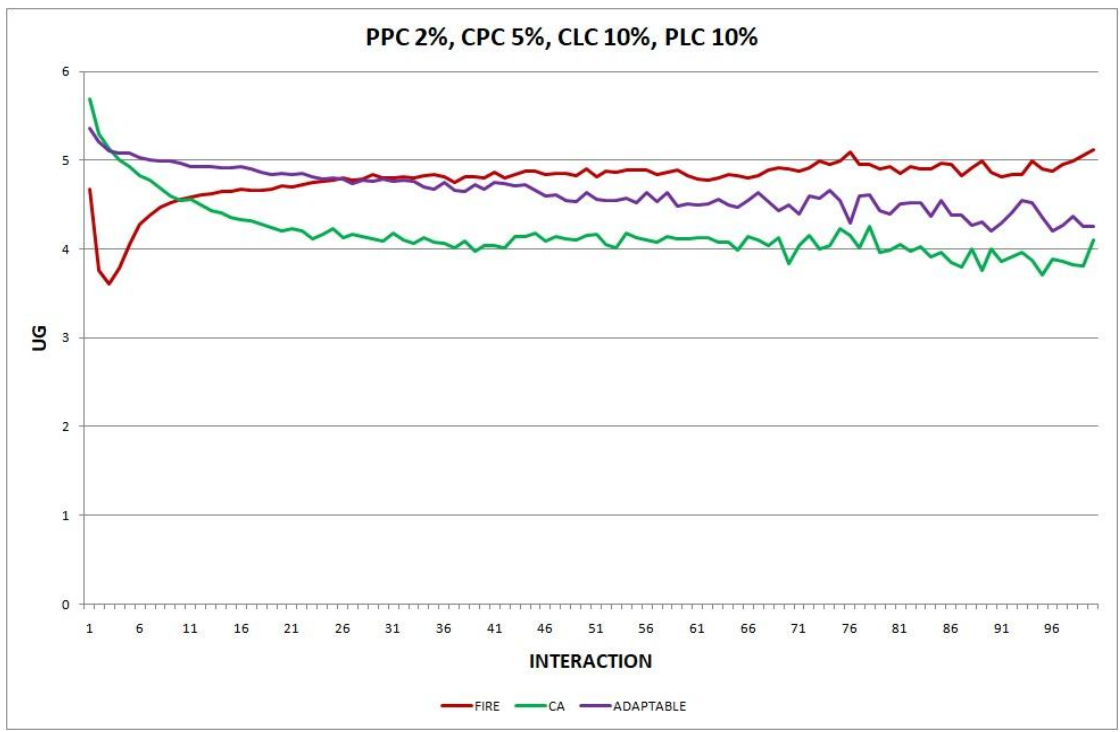

Fig. 15 Experiment 15: $p_{PPC} = 2\%, \ p_{CPC} = 5\%, \ p_{CLC} = 10\%, p_{PLC} = 10\%, \ \Delta\Phi = \pi/20$

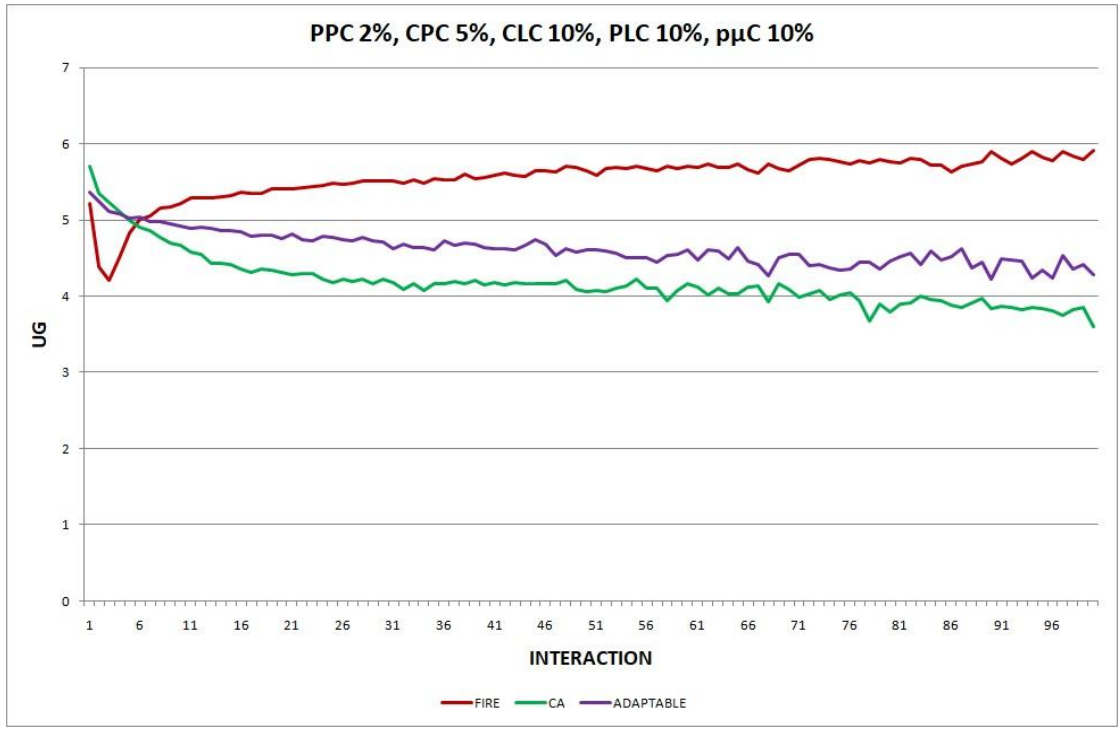

Fig. 16 Experiment 16: $p_{PPC} = 2\%, \ p_{CPC} = 5\%, \ p_{CLC} = 10\%, p_{PLC} = 10\%, \ \Delta\Phi = \pi/20, p_{\mu C} = 10\%, M = 1.0$

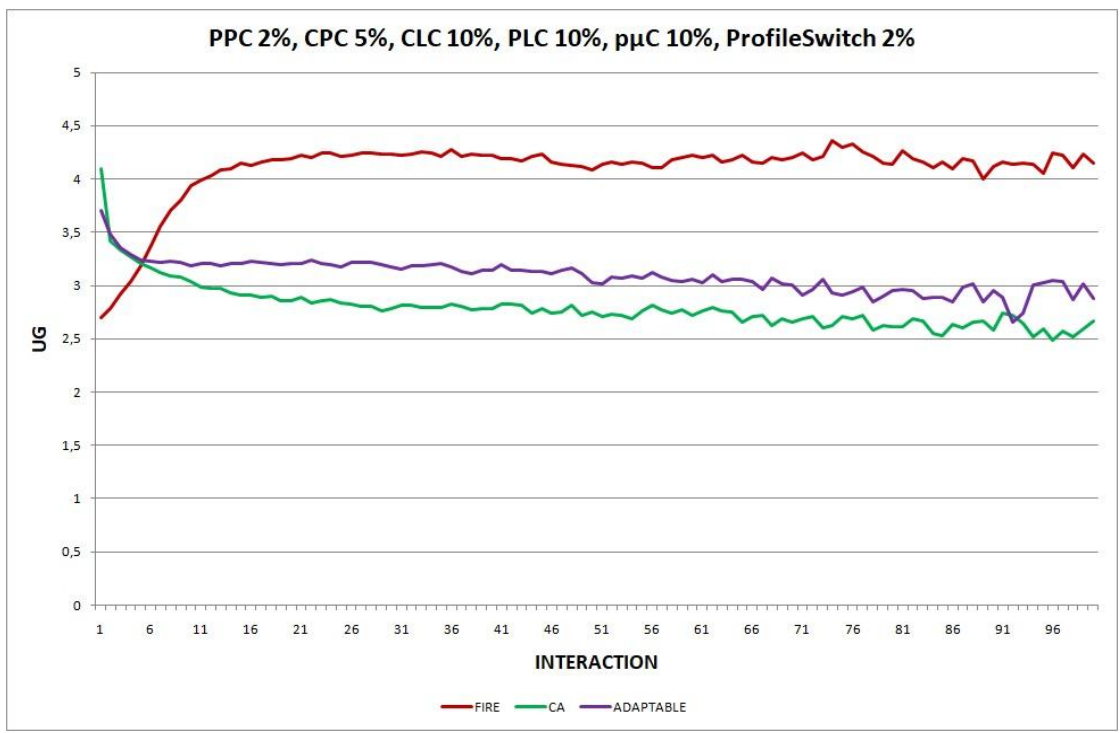

Fig. 17 Experiment 17: $p_{PPC} = 2\%, \ p_{CPC} = 5\%, \ p_{CLC} = 10\%, p_{PLC} = 10\%, \ \Delta\Phi = \pi/20, p_{\mu C} = 10\%, M = 1.0, p_{ProfileSwitch} = 2\%$

### 5.3. Adaptable's performance in changes that vary during simulation

In this section, we report the results of one final experiment, in which the environmental changes vary during simulation, which makes it even more difficult to predict which model

(FIRE or CA) is the best choice. The environmental changes applied in each simulation round are shown in Table 9.

**Table 9 Environmental changes applied in each simulation round. PPC stands for Provider Population Change, and CPC stands for Consumer Population Change.**

| Simulation round | Changes applied |
|---|---|
| 1-200 | PPC 2%, CPC 5% |
| 201-250 | PPC 2% |
| 251-300 | PPC 5% |
| 301-350 | PPC 10% |
| 351-400 | CPC 2% |
| 401-450 | CPC 5% |
| 451-500 | CPC 10% |

The experiment's results shown in figure Fig. 18 show that the adaptable consumers manage to find the optimal policy. During the initial interactions, adaptable consumers clearly choose CA as the best option, but when CA shows its worst performance, the adaptable consumers manage to keep their performance at higher levels. During the last interactions, when CA regains its good performance, adaptable consumers again adopt CA as the optimal choice.

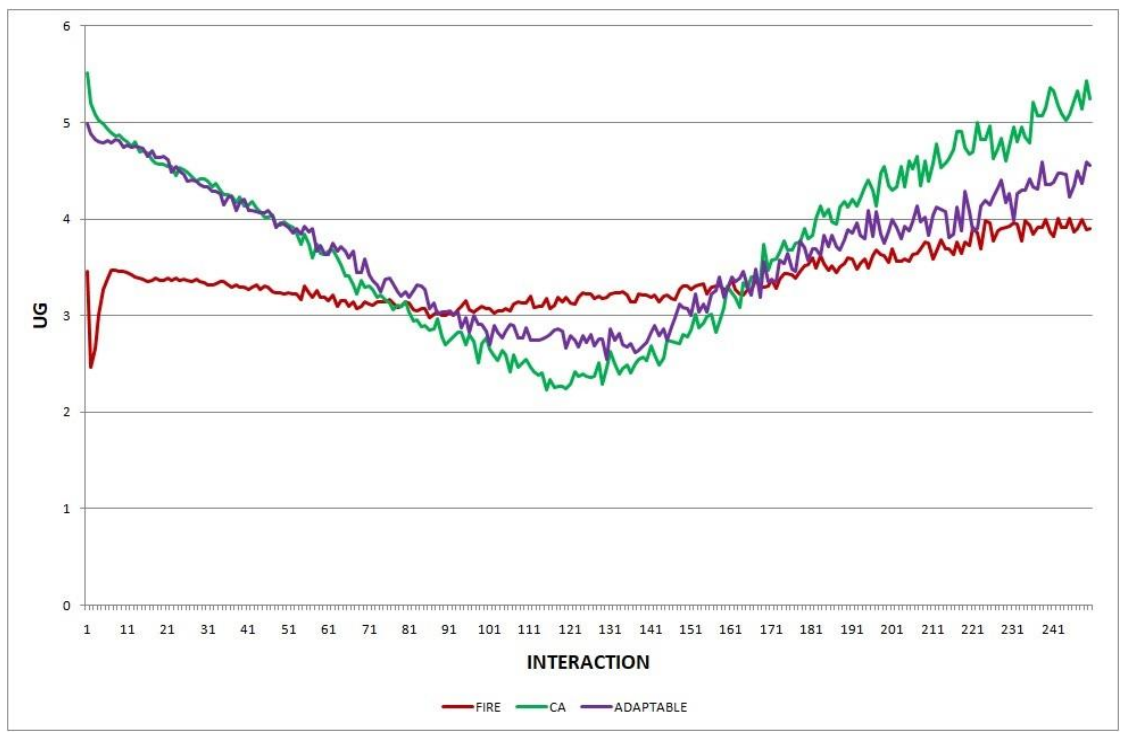

**Fig. 18 Experiment 18**

## 6. Discussion

In the majority of the experiments reported in section 5, the group of adaptable consumers manage to maintain performance levels between those of FIRE and CA consumer groups. This is to be expected, of course; learning to adapt in an environment is suggestive of performance that attempts, but cannot quite manage, to match the best alternative. However, there are a few experiments, in which the adaptable consumer outperforms the best alternative. We now discuss possible reasons for these findings.

In experiments 12 and 14, where the adaptable consumer outperforms the non-adaptable alternatives, there are two types of change which are in effect simultaneously (PPC 2%, CPC 5%). In the first type of change, the providers' population change (PPC), FIRE has the advantage, while in the second change, the consumers' population change (CPC), CA prevails. However, locally there are varying degrees of changes in several neighborhoods of agents. For example, some consumers may be newcomers in the system, while others are not, or new providers may exist in the operational range of only a few consumers. Unlike conventional, non-adaptable, consumers who exclusively apply one of the two models (FIRE or CA), adaptable consumers can detect locally which of the two types of behavior may be more warranted, try it and, maybe, revert back to the original one, thus demonstrating a greater agility to exploit the environmental conditions, leading to the superiority of the adaptable consumer in experiments 12 and 14.

What is the reason for the adaptable superiority in experiment 12, but not in experiment 13, where the change becomes more severe rising to 10%? We do not yet have a clear answer to this question, which hopefully will be the subject of future work. Nevertheless, adaptability is a property of the consumers; the longer they stay in the system, the more they learn to adapt. In experiment 13, increasing the rate of change of the consumer population (CPC) to 10%, has a negative effect on the performance of the adaptable consumers, as the newcomer adaptable consumers must learn to adapt from scratch and an increased rate of change does not allow for successful learned behaviors to be applied long enough to raise performance.

## 7. Conclusions and future work

Current trust and reputation models continue to have several unresolved issues, such as the inability to cope with agents' frequent entries and exits, as well as constantly changing behaviors. CA is a novel trust model, from the trustee's perspective, which aims to address these problems.

Previous research comparing CA to FIRE, an established trust model, found that CA outperforms in consumer population changes, whereas FIRE is more robust to provider population changes. The purpose of this research work is to investigate how to create an adaptable consumer agent, capable of learning when to use each model (FIRE or CA), so as to acquire maximum utility. This problem is framed as a reinforcement learning problem in a partially observable environment, in which the learning agent is unaware of the current state, able to learn through utility gained for choosing a trust model. We describe how the adaptable agent can measure a few features so as assess the current state of the environment and then use Deep Q learning to learn when to use the most profitable trust model. Our simulation experiments demonstrate that the adaptable consumer is able of finding the optimal policy in several simulated environmental conditions.

The ability to keep performance above that of the worst-choice model indicates the adaptable consumer's ability to learn the optimal policy. However, its performance usually falls below that of the best-choice model. This makes sense if we consider that the adaptable consumer chooses an action (push or pull) using an $\epsilon - greedy$ policy, a simple method to balance exploration and exploitation. The exploration-exploitation dilemma is central to Reinforcement Learning problems. Early in training, an agent has not learned anything meaningful in terms of associating higher Q-values with specific actions in different states, primarily due to lack of experience. Later on, once adequate experience has been accumulated, it should begin exploiting its knowledge to act optimally in the environment. The $\epsilon - greedy$ policy is a policy that chooses the best action (exploits) with probability $1 - \epsilon$, and a random action with probability $\epsilon$. In our experiments, we use a constant $\epsilon = 0.05$. It might be the case that the adaptable consumer explores too much and using a strategy for decaying epsilon would result in better utility gain for the adaptable consumer. We reserve investigating this as a future work.

Unwillingness and dishonesty in reporting of trust information is a persistent problem in agent societies. Yet, in the CA approach, agents do not share trust information, creating the expectation that CA is immune to various kinds of disinformation. Future research could look into how a trustor detects trust disinformation and learns to use the most appropriate trust model.

Ideally, a best performing machine learning model, i.e. its parameters and architecture, should be determined automatically by using a hyper-parameter tuning process. We only conducted an informal search due to the anticipated high computational cost. Computing the most accurate model on a node with limited resources may be impractical. Finding acceptable trade-offs between resource consumption and model accuracy to allow the deployment of a machine learning model capable of selecting the most suitable trust model in a resource-constrained environment is another interesting area of future research. In resource-constrained environments, being an adaptable agent using deep Q learning to select the optimal trust mechanism may be prohibitively expensive and should be an informed weighted choice.